%% file: mm_matryoshka.tex
\pdfoutput=1

\documentclass[11pt]{article}

\usepackage[final]{acl}
\usepackage{times}
\usepackage{latexsym}
\usepackage[T1]{fontenc}
\usepackage[utf8]{inputenc}
\usepackage{microtype}
\usepackage{inconsolata}
\usepackage{amsmath}
\usepackage{amsfonts}
\usepackage{amssymb}
\usepackage{graphicx}
\usepackage{booktabs}
\usepackage{multirow}
\usepackage{xcolor}
\usepackage{hyperref}
\usepackage{graphicx}
\usepackage{xspace}
\usepackage{url}
\newcommand{\method}{\textsc{MM-Matryoshka}\xspace}
\input{packages}

\title{\method: Towards Budget-Elastic Visual Document Retrieval \\via a 2D Multimodal Matryoshka Training Framework}

\author{\textbf{Haowen Xiang}$^{1}$, 
    \textbf{Yibo Yan}$^{1,2,3}$, 
    \textbf{Jiahao Huo}$^{1,2}$, 
    \textbf{Yu Huang}$^{1,2}$, 
    \textbf{Yi Cao}$^{2}$, 
    \textbf{Mingdong Ou}$^{2}$, 
    \textbf{Xuming Hu}$^{1,3}$\thanks{Corresponding Author}\\
    $^1$Hong Kong University of Science and Technology (Guangzhou), \\
    $^2$Alibaba Cloud Computing, 
    $^3$Hong Kong University of Science and Technology\\
    \texttt{\href{mailto:hoownn332@gmail.com}{hoownn332@gmail.com}},   \texttt{\href{mailto:xuminghu@hkust-gz.edu.cn}{xuminghu@hkust-gz.edu.cn}}
    \vspace{-3mm}}

\begin{document}
\maketitle

\begin{abstract}
Multi-vector visual document retrievers achieve strong fine-grained matching by representing each page with multiple vectors from deep Vision-Language Models (VLMs), but this design \textit{makes deployment expensive in both storage and computational overhead}. Existing efficiency techniques usually optimize only part of this budget, leaving multimodal retrievers \textit{without a unified way to trade accuracy for both vector width and encoder depth}. Therefore, we propose \textbf{\method, a 2D Matryoshka training framework for budget-elastic Visual Document Retrieval (VDR)}, enabling ColPali-style multi-vector retrieval elastic along both dimension and layer. At inference time, a single retriever can select a 2D selectable budget without training separate models for different budgets. Through comprehensive experiments across multiple representative backbones, we demonstrate that by retaining significantly higher quality than direct truncation baselines while substantially reducing storage and computational overhead, \method can offer robust budget elasticity for efficient VDR.

\end{abstract}

\section{Introduction}
\label{sec:intro}
VDR aims to retrieve relevant document pages given natural-language queries while preserving evidence from text, layout, tables, charts, and visual regions \cite{yan2026unlocking}. Recent ColPali-style systems~\citep{faysse2025colpali} adapt VLMs to this setting with late interaction~\citep{khattab2020colbert}: queries and pages are encoded into multiple vectors, and relevance is computed by matching each query vector to its most similar page vector. Though effective, such a design is costly in deployment, as the machine must store many vectors. \texttt{MaxSim} operation scales with the number of query vectors, page vectors, and embedding dimensions, and modern VLM-based encoders are deep. A practical VDR retriever therefore needs not only high full-budget accuracy, but also a controllable budget that trades accuracy for storage and latency.

\begin{figure}[t!]
    \centering
    \includegraphics[width=1 \linewidth]{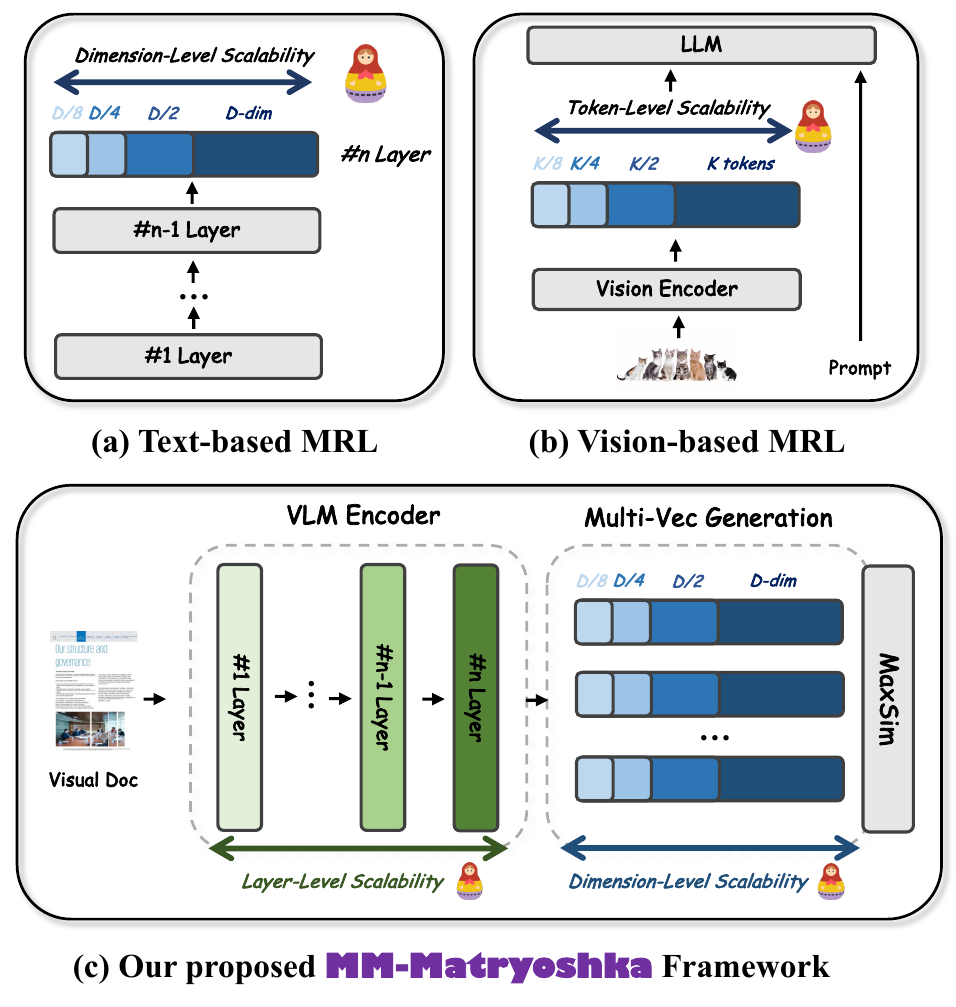}
    \vspace{-2em}
    \caption{Comparison between previous text-based (a) \& vision-based MRL (b) vs. \method (c).}
    \label{fig:mrl_paradigm_comparison}
    \vspace{-6mm}
\end{figure}

One of the efficient embedding techniques is \textbf{Matryoshka Representation Learning (MRL)}, which trains a single embedding to expose nested prefix dimensions, so that shorter prefixes retain useful semantic information~\citep{kusupati2022matryoshka}, as shown in Figure \ref{fig:mrl_paradigm_comparison} (a). Recent work shows that dimension-nested embeddings can reduce index storage and dot-product cost while using a single trained model~\citep{yoon2024matryoshka,zhang2025smec,xiao2025metaembed}. This line of work establishes an important idea: a single backbone can be trained to yield selectable test-time budgets, rather than requiring separate models. However, the conventional MRL setting \textit{mainly controls vector width, leaving encoder depth outside the budget space}.

In the past year, the MRL principle has begun to enter the multimodal domain.  As illustrated in Figure \ref{fig:mrl_paradigm_comparison} (b), Matryoshka Multimodal Models~\citep{cai2025matryoshka} learn nested sets of multiple visual tokens that represent visual content at multiple coarse-to-fine granularities, enabling test-time control over how much visual detail is passed to the LLM backbone. Related multimodal retrieval work also explores flexible interaction budgets for multimodal retrieval~\citep{xiao2025metaembed}. This direction is especially relevant for VDR because multi-vector pages multiply the storage of every retained dimension or token. Yet existing multimodal MRL-related compression still leaves a major source of cost untouched: the retriever \textit{typically relies on final-layer representations from a VLM encoder}. 

Therefore, we propose \textbf{\method, a 2D matryoshka training framework, which aims to make both the vector width and the encoder depth budget elastic in the VDR domain}, as illustrated in Figure \ref{fig:mrl_paradigm_comparison} (c).
Specifically, dimension-level MRL loss makes multiple vector prefixes directly retrievable, and layer-level MRL loss transfers final-layer retrieval behavior to sampled intermediate layers. At inference time, the same model selects a layer $\ell$ and prefix dimension $d$, normalizes the selected vector prefixes, and scores with the same \texttt{MaxSim} function. This design jointly targets storage and encoder latency through $d$ and $\ell$, while retaining the fine-grained matching behavior that makes VDR accurate. Across representative VLM backbones and 14 VDR benchmark \cite{mace2025vidore,faysse2025colpali}, our results show robust budget-elastic behavior of \method.

Our contributions can be summarized as follows:
\begin{itemize}[leftmargin=*]
    \item[\ding{182}] We are the first to formulate a 2D training framework \method for VDR, enabling a single retriever to balance efficiency and accuracy across both vector width and encoder depth.
    
    \item[\ding{183}] \method is robust across representative VLM backbones, achieving the expected budget-elastic behavior through improved low-dimensional and intermediate-layer retention.
    
    \item[\ding{184}] We conduct a comprehensive experimental validation across 14 VDR datasets, covering both performance and efficiency, and complement the quantitative results with qualitative case studies.
\end{itemize}

\section{Related Work}
\label{sec:related}

\subsection{Visual Document Retrieval}

VDR retrieves document pages from the multimodal corpus rather than from extracted text alone \cite{osmulski2025miracl,yan2026beyond}. ColPali~\citep{faysse2025colpali} showed that VLMs can be adapted to this setting with late interaction, making it possible to match query vectors against fine-grained page regions. 
ViDoRe team~\citep{mace2025vidore,faysse2025colpali} further provides benchmarks for evaluating visual retrieval across diverse document domains. In parallel, universal multimodal embedding models have advanced VDR with \textbf{single-vector} representations~\citep{jiang2024vlm2vec,zhang2024gme,chen2025mme5,gunther2025jina,xu2025llama}. 
Recent \textbf{multi-vector} systems such as Llama-Nemoretriever-Colembed \cite{xu2025llama} and Nemotron ColEmbed V2 \cite{moreira2026nemotron} maintain top-tier accuracy but severely exacerbate storage bottlenecks by demanding thousands of high-dimensional embeddings per document. 
\method focuses on a critical efficiency question in the VDR domain: \textit{for multi-vector VDR systems that intentionally keep many vectors, can the same retriever expose a lower-cost selectable budget?}

\subsection{Efficient Multi-Vector Retrieval}

Efficient multi-vector retrieval aims to preserve the accuracy benefits of late interaction while mitigating its prohibitive storage footprint and search latency \cite{lee2023rethinking,archish2026incorporating,shrestha2024espn}.
In the \textbf{textual domain}, foundational works like ColBERTv2 \cite{santhanam2022colbertv2} and PLAID \cite{santhanam2022plaid} utilize residual compression and centroid-based pruning, while CITADEL \cite{li2023citadel} routes token vectors through dynamic lexical keys.
Recently, the severe memory bottleneck of VLMs has spurred a wave of efficiency optimizations specifically tailored for VDR.
These \textbf{multimodal efforts} primarily focus on token reduction along the sequence dimension: pruning-based methods like DocPruner \cite{yan2025docpruner,yan2026sculpting,bach2025hierarchical} adaptively discard redundant visual patches based on attention scores, whereas merging-based frameworks such as Light-ColPali \cite{ma2025towards} and ColChunk \cite{yan2026visual} aggregate dense patch embeddings via semantic or layout-aware clustering.
 Furthermore, advanced hybrid approaches attempt to compress indices through learnable summary tokens or attention-guided grouping \cite{xiao2025metaembed,qin2026multi,kim2025hybrid,huo2026causalembed}.
 While these approaches successfully reduce the token count or optimize the retrieval engine, they still \textit{operate at a fixed representation scale}. \method offers a fundamentally distinct and complementary paradigm: rather than solely optimizing token quantity, we \textit{empower a single retriever to be naturally budget-elastic along both vector width and encoder layer depth}.

\begin{figure*}[htb!]
  \centering
  \includegraphics[width=\textwidth]{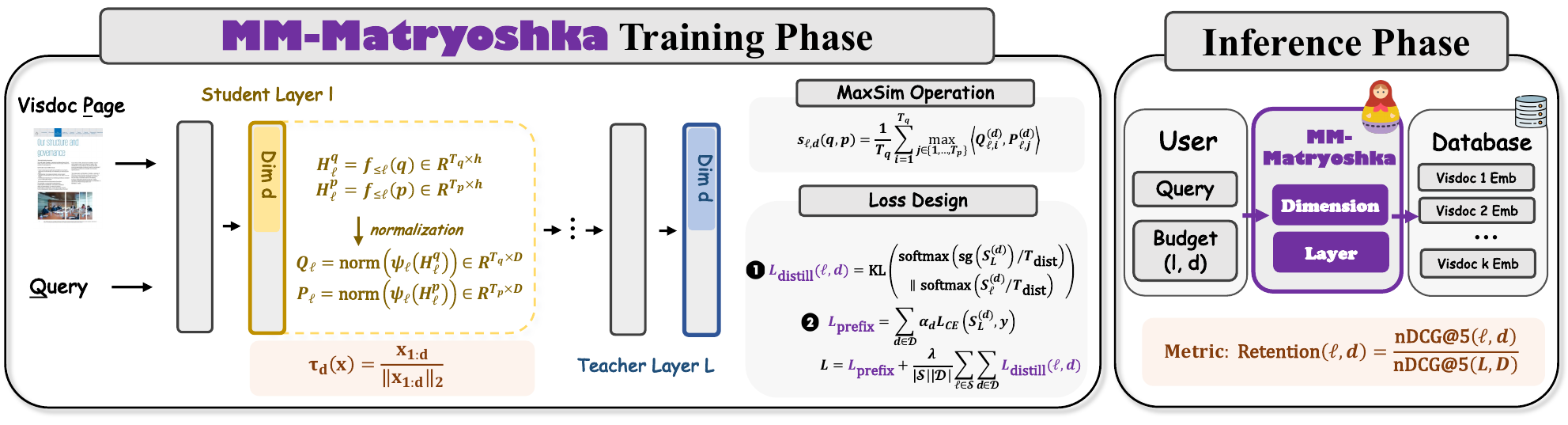}
  \caption{The overall training and inference framework of \method.}
  \label{fig:mm_matryoshka_framework}
\end{figure*}

\subsection{Matryoshka Representation Learning}

MRL trains embeddings whose prefix dimensions remain useful under different computational budgets~\citep{kusupati2022matryoshka}. Recent work adapts this idea to existing LLM embeddings and retrieval-oriented embedding compression~\citep{yoon2024matryoshka,zhang2025smec}. More closely related to our budget formulation, 2D Matryoshka methods expose a joint space over embedding width and Transformer layer depth for sentence embeddings~\citep{li20242d,wang20252d,zhuang2024starbucks}. MRL has also been explored in VLMs by controlling visual-token granularity rather than retrieval-vector width or encoder depth~\citep{cai2025matryoshka}. However, most existing studies are based on text retrieval or general representation learning \cite{lai2024matryoshkarec,wang2024train,huy2026mipic,wang2026m3bert,bussmann2025learning}. Multi-vector VDR introduces a different constraint: compression must preserve many vectors-to-region alignments used by late interaction \cite{faysse2025colpali}. We therefore investigate 2D Matryoshka training specifically in VDR, where the goal is not only smaller embeddings but a 2D budget space.

\section{Methodology}
\label{sec:method}

We adapt the MRL idea from single-vector sentence embeddings~\citep{li20242d,wang20252d,zhuang2024starbucks} to multi-vector VDR. Instead of producing one sentence vector $\mathbf{X}_{n}^{d}$ from layer $n$ and dimension $d$, a VDR model produces many query and page token vectors, and retrieval quality is determined by a late-interaction score matrix. The goal of \method is therefore to make the score matrix reliable under a selected lower encoder layer $\ell$ and a reduced prefix dimension $d$, as illustrated in Figure \ref{fig:mm_matryoshka_framework}.

\subsection{Problem Setup and Budget Space}

Given a batch of paired text queries and rendered document pages $\{(q_i,p_i)\}_{i=1}^{B}$, where $B$ is the batch size and $(q_i,p_i)$ is the $i$-th query--page pair, let $f_{\leq \ell}$ denote the VLM encoder truncated after transformer layer $\ell$. For a selected layer $\ell$, the encoder first produces query and page hidden states:

\begin{equation}
\begin{split}
    \mathbf{H}_{\ell}^{q} &= f_{\leq \ell}(q) \in \mathbb{R}^{T_q\times h}, \\
    \mathbf{H}_{\ell}^{p} &= f_{\leq \ell}(p) \in \mathbb{R}^{T_p\times h},
\end{split}
\end{equation}
where $\mathbf{H}$ denotes an encoder hidden-state matrix: $\mathbf{H}_{\ell}^{q}$ and $\mathbf{H}_{\ell}^{p}$ are the query and page hidden-state matrices at layer $\ell$, $T_q$ and $T_p$ are the numbers of query and page vectors, and $h$ is the pre-projection, backbone-native hidden-state dimension. We then apply a layer-specific retrieval projection followed by per-token normalization to obtain $D$-dimensional retrieval vectors:

\begin{equation}
\begin{split}
    \mathbf{Q}_{\ell} &= \mathrm{norm}\!\left(\psi_{\ell}(\mathbf{H}_{\ell}^{q})\right) \in \mathbb{R}^{T_q\times D}, \\
    \mathbf{P}_{\ell} &= \mathrm{norm}\!\left(\psi_{\ell}(\mathbf{H}_{\ell}^{p})\right) \in \mathbb{R}^{T_p\times D},
\end{split}
\end{equation}

where $\psi_{\ell}$ is the projection head for layer $\ell$, and $\mathrm{norm}(\cdot)$ applies $\ell_2$ normalization independently to each token vector. The final layer $L$ gives the standard full-depth retriever, while $\ell<L$ gives a shallower selectable budget.

For a prefix dimension $d\leq D$, we follow Matryoshka prefix truncation and define the normalized prefix of a token vector $x\in\mathbb{R}^{D}$ as
\begin{equation}
    \tau_d(\mathbf{x}) =
    \frac{\mathbf{x}_{1:d}}
         {\lVert \mathbf{x}_{1:d}\rVert_2}.
\end{equation}
We apply $\tau_d$ to each query and page token vector. The resulting budgeted representations are $\mathbf{Q}_{\ell}^{(d)}=\tau_d(\mathbf{Q}_{\ell})$ and $\mathbf{P}_{\ell}^{(d)}=\tau_d(\mathbf{P}_{\ell})$. A 2D budget is therefore a pair $(\ell,d)$.

Under budget $(\ell,d)$, the relevance score between query $q$ and page $p$ is computed with the same \texttt{MaxSim} operator:
\begin{equation}
    s_{\ell,d}(q,p)=
    \frac{1}{T_q}
    \sum_{i=1}^{T_q}
    \max_{j\in\{1,\ldots,T_p\}}
    \left\langle
    \mathbf{Q}_{\ell,i}^{(d)}, \mathbf{P}_{\ell,j}^{(d)}
    \right\rangle .
\end{equation}
For a training batch, we denote by $S_{\ell}^{(d)}\in\mathbb{R}^{B\times B}$ the in-batch score matrix whose $(i,j)$ entry is $s_{\ell,d}(q_i,p_j)$.

% \subsection{Prefix-Dimensional Late-Interaction Loss}
\subsection{Dimension-Level MRL}

The first training objective makes the final-layer retriever usable at multiple embedding widths. For a set of prefix dimensions $\mathcal{D}$, we compute a late-interaction score matrix at each prefix and apply the in-batch retrieval loss:
\begin{equation}
    \mathcal{L}_{\mathrm{prefix}}
    =
    \sum_{d\in\mathcal{D}}
    \alpha_d
    \mathcal{L}_{\mathrm{CE}}(\mathbf{S}_{L}^{(d)}, \mathbf{y}),
\end{equation}
where $\mathbf{S}_{L}^{(d)}$ is the final-layer in-batch score matrix at prefix dimension $d$, $\mathbf{y}$ is the label vector with $y_i=i$ so that the paired page is the positive for query $q_i$, $\mathcal{L}_{\mathrm{CE}}$ is the cross-entropy loss, and $\alpha_d$ weights the contribution of prefix $d$. This objective trains every selected prefix through the same late-interaction scoring function used at inference.

% \subsection{Layer Distillation}
\subsection{Layer-Level MRL}

The second objective extends the budget from vector width to encoder depth. We define a target layer set $\mathcal{S}\subset\{1,\ldots,L-1\}$, where $\mathcal{S}$ contains intermediate layers trained as student layers. For each training batch, we enumerate the full grid of target layers and prefix dimensions: for every $\ell\in\mathcal{S}$ and $d\in\mathcal{D}$, we compute the final-layer score matrix $\mathbf{S}_{L}^{(d)}$ and the target-layer score matrix $\mathbf{S}_{\ell}^{(d)}$ under the same prefix dimension $d$.

We distill the final-layer in-batch score distribution to each intermediate layer. The final layer serves as the teacher and each target layer serves as a student:
\begin{equation}
\begin{split}
    \mathcal{L}_{\mathrm{distill}}(\ell,d) = \mathrm{KL} \Big( 
        &\mathrm{softmax}(\mathrm{sg}(\mathbf{S}_{L}^{(d)}) / T_{\mathrm{dist}}) \\
        &\parallel \mathrm{softmax}(\mathbf{S}_{\ell}^{(d)} / T_{\mathrm{dist}}) 
    \Big),
\end{split}
\end{equation}

where $\mathrm{KL}$ denotes Kullback--Leibler divergence between in-batch score distributions computed separately for each query over its $B$ candidate pages, $T_{\mathrm{dist}}$ is the distillation temperature, and $\mathrm{sg}(\cdot)$ stops gradients through the teacher scores. This score-level alignment is important for VDR because many different token-vector configurations can yield similar \texttt{MaxSim} matches. Matching score distributions therefore encourages the intermediate layer to preserve retrieval decisions without over-constraining its token-level geometry.

\subsection{Joint Loss Design and Inference}

% Figure~\ref{fig:mm_matryoshka_framework} summarizes the complete training and inference procedure. 
The full training objective combines final-layer prefix supervision with score distillation over all selected 2D pairs in $\mathcal{S}\times\mathcal{D}$:

\begin{equation}
    \mathcal{L}
    =
    \mathcal{L}_{\mathrm{prefix}}
    +
    \frac{\lambda}{|\mathcal{S}|\,|\mathcal{D}|}
    \sum_{\ell\in\mathcal{S}}
    \sum_{d\in\mathcal{D}}
    \mathcal{L}_{\mathrm{distill}}(\ell,d),
\end{equation}
where $\lambda$ controls the strength of layer-wise distillation, and the normalization averages the distillation loss over the selected 2D grid $\mathcal{S}\times\mathcal{D}$. The target layer set $\mathcal{S}$ and prefix set $\mathcal{D}$ are chosen according to the characteristics of each backbone, including its encoder depth, retrieval-head width, and the availability of useful intermediate hidden states. This backbone-dependent budget design lets the same objective cover different 2D trade-offs.

At inference time, no separate retriever is trained for a new budget. As shown in inference phase of Figure~\ref{fig:mm_matryoshka_framework}, we choose a 2D pair $(\ell,d)$, extract the corresponding query and page token vectors, apply prefix truncation and re-normalization, and score with the same \texttt{MaxSim} function. Dimension reduction directly reduces token-vector storage and dot-product width. Layer reduction uses hidden states from an earlier layer; with an early-exit implementation, it can also reduce encoder computation.

\section{Experiments}
\label{sec:experiments}

\subsection{Experimental Setup}

\subsubsection{Datasets and Metrics}

All trainable variants are trained on \texttt{vidore/colpali\_train\_set}\footnote{\url{https://huggingface.co/datasets/vidore/colpali_train_set}} and evaluated under the ViDoRe protocol~\citep{faysse2025colpali,mace2025vidore}. We evaluate on two ViDoRe benchmark suites: ViDoRe v1\footnote{\url{https://huggingface.co/collections/vidore/vidore-benchmark}} and ViDoRe v2\footnote{\url{https://huggingface.co/collections/vidore/vidore-benchmark-v2}}. Unless otherwise stated, results are averaged separately over the 10 sub-datasets in ViDoRe v1 and the 4 sub-datasets in ViDoRe v2. Following the benchmark protocol, we use nDCG@5 as the primary metric and also compute Recall@1 and Recall@5. To evaluate compressed budgets, we additionally report absolute nDCG@5 and retention relative to each model's final-layer full-width reference setting:
\begin{equation}
    \mathrm{Retention}(\ell,d)
    =
    \frac{\mathrm{nDCG@5}(\ell,d)}
         {\mathrm{nDCG@5}(L,D)}.
\end{equation}
All nDCG@5 values in the tables are multiplied by 100 for readability, and retention values are reported as percentages.

\subsubsection{Backbones and Baselines}

We trained and evaluated three vision-language backbone families to test whether 2D budget elasticity is architecture-specific or broadly reproducible: PaliGemma~\citep{beyer2024paligemma}, Qwen2-VL~\citep{wang2024qwen2}, and Qwen2.5-VL~\citep{wu2025qwen}. Each trained variant is compared with the corresponding public ColPali-family late-interaction retriever~\citep{faysse2025colpali}: ColPali for PaliGemma, ColQwen2 for Qwen2-VL, and ColQwen2.5 for Qwen2.5-VL. Table~\ref{tab:model-details} in the appendix lists the model sizes and Hugging Face repos. We will release our codebase and checkpoints to the community.

\subsubsection{Efficiency-Oriented Budget Evaluation}

We evaluate efficiency as a quality--cost trade-off over selectable 2D budgets. For retrieval quality, each trained checkpoint is evaluated at budget $(\ell,d)$ by selecting the representation from a target layer $\ell$. We evaluate different target-layer sets depending on the backbone, together with the shared Matryoshka dimension set.

For measured deployment efficiency, we report a PaliGemma study on the same datasets using the same trained checkpoint. The measured selected grid combines the target layer and dimension set, using the final-layer full-width budget as the reference. We organize the efficiency protocol around three deployment costs. Indexing--storage measures the bf16 page-vector footprint after compression, computed as the number of page tokens times the selected dimension and two bytes per value. Indexing latency measures the page-encoding time needed to build the compressed embedding index, with early exit at the selected retrieval layer. Query latency measures query processing under the selected budget, including query encoding and a fixed number of candidate \texttt{MaxSim} scoring with precomputed page embeddings. \texttt{MaxSim} timing uses fixed sampled candidate sets, CUDA synchronization around each timed call, and includes host-to-device transfer of query embedding and host-side score transfer, while excluding disk I/O and first-stage retrieval.

\subsection{Main Results}
\label{sec:results}

We organize the main results around the budget-elastic behavior of \method. Section~\ref{sec:compression-retention} reports two-dimensional compression retention, showing how retrieval quality changes when both vector width and encoder depth are reduced. Section~\ref{sec:efficiency-evaluation} presents an efficiency evaluation that connects compressed budget to index storage, indexing latency, and query latency. Section~\ref{sec:case-study} provides a qualitative case that compares a highly compressed budget with the baseline. Detailed raw experimental results, including complete per-dataset all-budget reference scores and the raw efficiency grid, are provided in the appendix.

\subsubsection{2D Compression Retention}
\label{sec:compression-retention}

\begin{table*}[t]
\centering
\scriptsize
\setlength{\tabcolsep}{2.5pt}
\resizebox{\textwidth}{!}{%
\begin{tabular}{cl*{12}{r}}
\toprule
\multirow{3}{*}{Backbone} & \multirow{3}{*}{Model} &
\multicolumn{6}{c}{Middle layer} &
\multicolumn{6}{c}{Shallow layer} \\
\cmidrule(lr){3-8}\cmidrule(lr){9-14}
& & \multicolumn{2}{c}{$128$d} & \multicolumn{2}{c}{$64$d} & \multicolumn{2}{c}{$32$d} &
\multicolumn{2}{c}{$128$d} & \multicolumn{2}{c}{$64$d} & \multicolumn{2}{c}{$32$d} \\
\cmidrule(lr){3-4}\cmidrule(lr){5-6}\cmidrule(lr){7-8}
\cmidrule(lr){9-10}\cmidrule(lr){11-12}\cmidrule(lr){13-14}
& & v1 & v2 & v1 & v2 & v1 & v2 & v1 & v2 & v1 & v2 & v1 & v2 \\
\midrule
\multirow{2}{*}{PaliGemma} & ColPali & 53.8\% & 37.4\% & 37.3\% & 20.0\% & 13.4\% & 14.4\% & 42.8\% & 26.7\% & 26.0\% & 15.7\% & 11.2\% & 10.8\% \\
& + \method & 98.7\% & 91.0\% & 96.6\% & 86.9\% & 93.9\% & 74.3\% & 91.9\% & 80.3\% & 90.8\% & 75.5\% & 86.7\% & 63.9\% \\
\midrule
\multirow{2}{*}{Qwen2-VL} & ColQwen2 & 76.6\% & 52.3\% & 58.2\% & 27.7\% & 37.0\% & 20.9\% & 84.9\% & 67.5\% & 75.2\% & 49.6\% & 59.4\% & 34.2\% \\
& + \method  & 98.2\% & 93.6\% & 95.8\% & 86.4\% & 91.6\% & 78.1\% & 96.0\% & 83.3\% & 93.6\% & 80.7\% & 88.3\% & 72.4\% \\
\midrule
\multirow{2}{*}{Qwen2.5-VL} & ColQwen2.5 & 77.8\% & 60.0\% & 69.0\% & 49.7\% & 47.8\% & 31.4\% & 78.2\% & 53.6\% & 65.3\% & 40.5\% & 36.8\% & 20.6\% \\
& + \method  & 101.7\% & 100.6\% & 101.0\% & 96.0\% & 97.6\% & 85.8\% & 98.1\% & 90.6\% & 96.4\% & 85.3\% & 92.7\% & 72.9\% \\
\bottomrule
\end{tabular}
}
\caption{2D compression retention on ViDoRe v1 and v2 averages. Values are percentages relative to each model's highest-budget reference. For all rows, the reference is final-layer $128$d. Middle layer denotes $0.5L$ (PaliGemma layer 9, Qwen2-VL layer 14, Qwen2.5-VL layer 18); shallow layer denotes layers 4, 7, and 9, respectively.}
\label{tab:compression-retention}
\end{table*}

\begin{figure*}[t]
\centering
\includegraphics[width=0.95\textwidth]{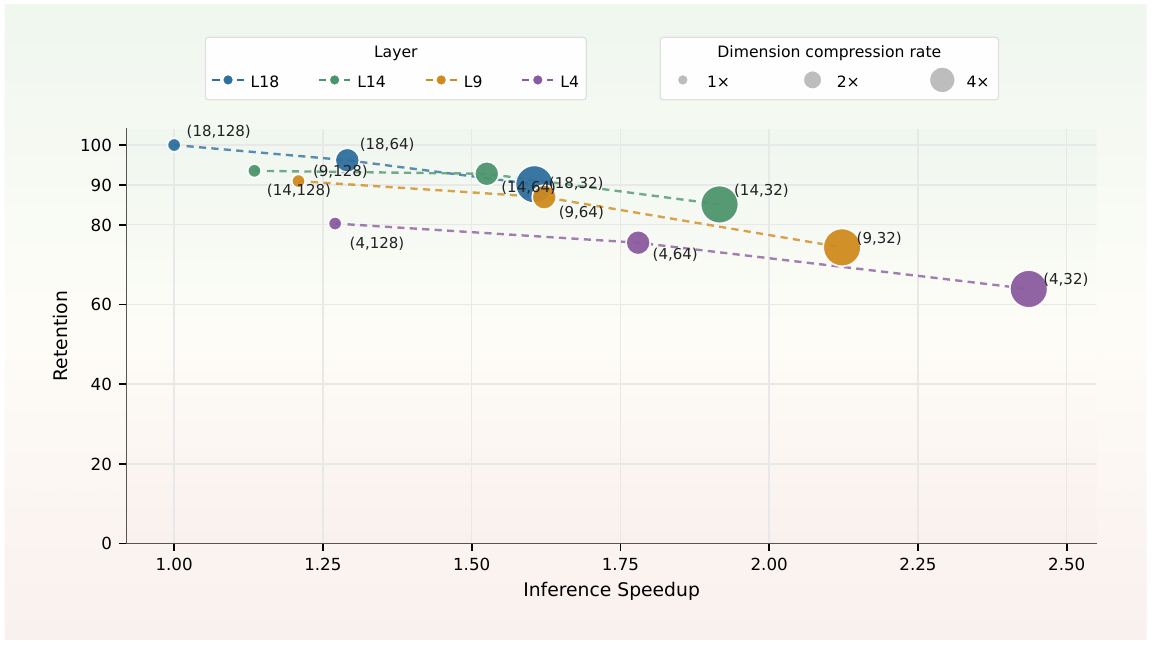}
\caption{Quality--efficiency trade-off of \textit{PaliGemma+\method} over the 15 2D budgets. Higher is better along both axes; the dashed line marks 80\% nDCG@5 retention, marker area indicates storage compression, and labels show $(\ell,d)$.}
\label{fig:paligemma-efficiency-pareto}
\end{figure*}

The main benefit of 2D Matryoshka training is that retrieval quality degrades gracefully when vector width and encoder depth are reduced together. Table~\ref{tab:compression-retention} reports a compact 2D retention grid on both ViDoRe v1 and ViDoRe v2. The table shows the compressed operating points: each row compares middle-layer and shallow-layer budgets at $128$, $64$, and $32$ dimensions against the same model's highest-budget final-layer setting. Starting from comparable full-budget retrievers, the consistent pattern is that the trained models preserve a much larger fraction of their own highest-budget quality than the baselines that have not been explicitly trained for these compressed operating points.

The key result is that the two compression axes can be imposed simultaneously: even under joint layer and dimension reduction, the trained retrievers preserve a large majority of their highest-budget quality and remain far ahead of the same compressed budgets applied to public baselines. At the middle layer with 32 dimensions, the trained models maintain a significant portion of their highest-budget quality across the v1/v2 averages, with retention rates reaching as high as $97.6\%$, while the corresponding baselines retain only $47.8\%$. At the more aggressive shallow-layer $32$-dimensional budget, the trained models still retain $63.9\%$--$92.7\%$, whereas the baselines fall to $10.8\%$--$59.4\%$. This large gap supports the intended 2D Matryoshka behavior of \method: prefix-dimensional training keeps narrow token vectors useful, and layer-wise distillation makes intermediate representations retrieval-ready under the same late-interaction scoring interface. The complete raw evaluation data is provided in Appendix~\ref{app:raw-compression}.

\subsubsection{Efficiency Evaluation}

\label{sec:efficiency-evaluation}

\begin{figure}[t]
\centering
\includegraphics[width=\linewidth]{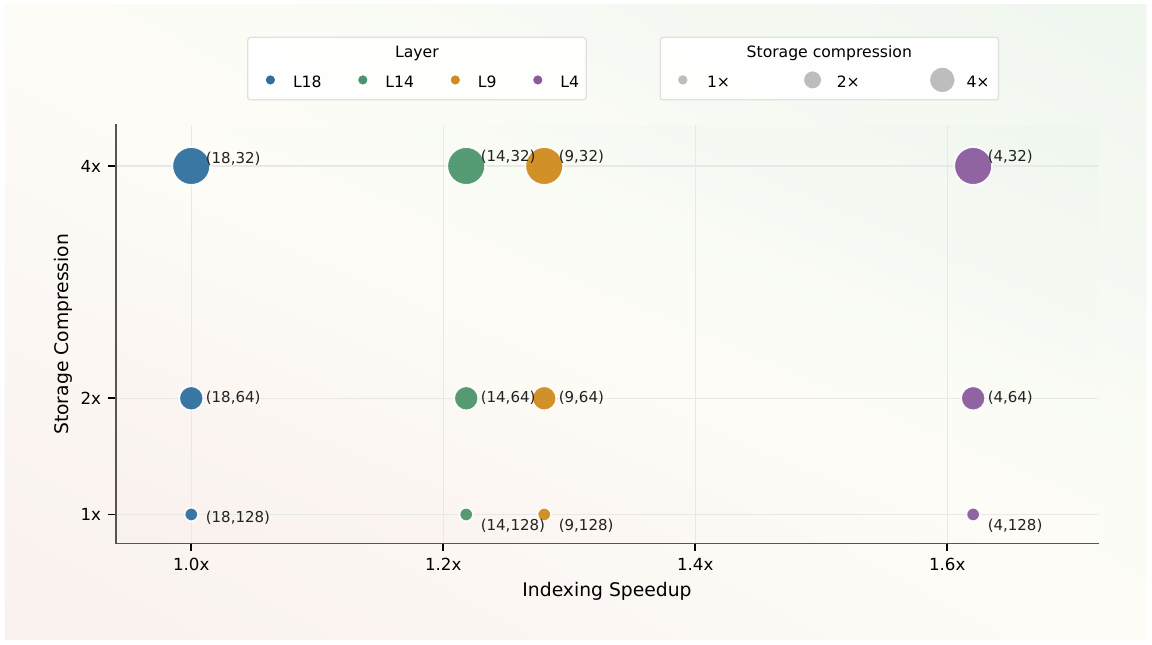}
\caption{Indexing--storage trade-off of \textit{PaliGemma+\method}. The $x$-axis reports page-indexing speedup relative to the full $L18$-$D128$ budget, and the $y$-axis reports embedding-index storage compression relative to $D128$; labels show $(\ell,d)$.}
\label{fig:paligemma-index-storage-pareto}
\end{figure}

The compression results translate into practical deployment choices for both inference and vector-index construction. Prior 2D Matryoshka sentence-embedding work observes that Transformer inference time grows approximately linearly with the number of executed layers~\citep{li20242d}. In our setting, layer reduction affects encoder cost, while dimension reduction affects both storage and \texttt{MaxSim} dot-product width. We therefore report the PaliGemma efficiency study from two complementary views. First, Figure~\ref{fig:paligemma-efficiency-pareto} pairs the retrieval-quality grid from the four-dataset ViDoRe v2 average with inference speed: the $x$-axis reports speedup relative to the full $L18$-$D128$ budget, and the $y$-axis reports nDCG@5 retention relative to the same reference. Second, Figure~\ref{fig:paligemma-index-storage-pareto} shows how shallower page encoding reduces indexing time while lower dimensions reduce embedding-index storage. In both figures, each label gives the selected $(\ell,d)$ budget, where $\ell$ is the layer and $d$ is the dimension. See the full measurement grid in Appendix~\ref{app:efficiency}.

Together, Figures~\ref{fig:paligemma-efficiency-pareto} and~\ref{fig:paligemma-index-storage-pareto} show that \method exposes a range of selectable budgets rather than a single compressed model. For inference, final-layer width compression gives a conservative trade-off: $(18,32)$ keeps $90.1\%$ of the full-budget nDCG@5 while reducing the embedding index to one quarter of its original size and improving inference latency by $1.78\times$. Joint 2D budgets provide more aggressive budgets. For example, $(9,32)$ keeps $74.3\%$ of the full-budget quality with the same $4\times$ storage compression and a $2.17\times$ inference speedup, while $(4,128)$ stays near the 80\% retention line and mainly trades layer depth for indexing and query-encoding speed. This behavior matches the use of \method: deployment can select a budget according to a target quality floor, storage limit, or latency target, while using the same retriever and \texttt{MaxSim} interface.

\subsubsection{Qualitative Cases}
\label{sec:case-study}

\begin{figure*}[t]
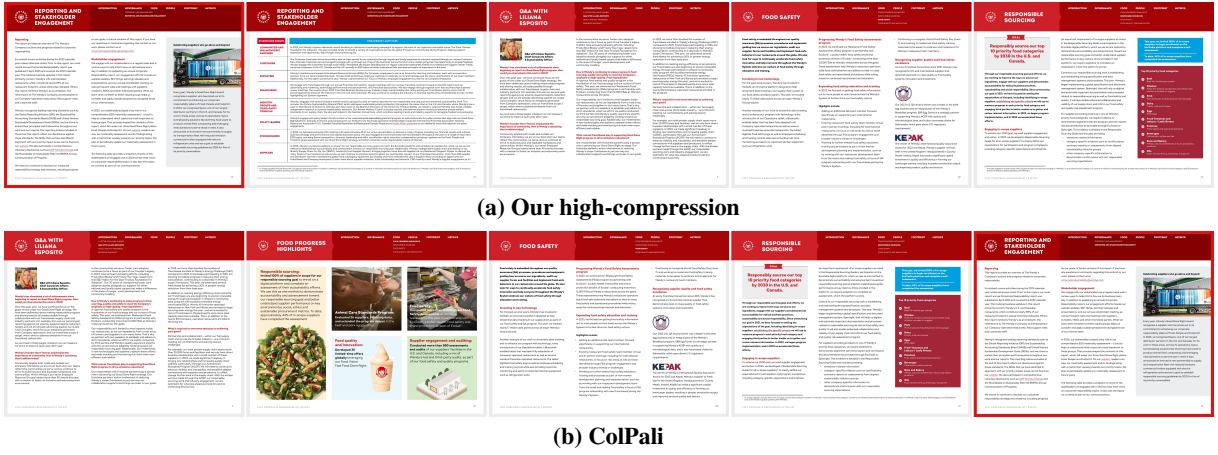

\centering
\includegraphics[width=\textwidth]{figure/top5_montage_paper_clean_ours.png}\vspace{-0.3em}
\\{\footnotesize\textbf{(a) Our high-compression}}

\vspace{0.3em}

\includegraphics[width=\textwidth]{figure/top5_montage_paper_clean_base.png}\vspace{-0.3em}
\\{\footnotesize\textbf{(b) ColPali}}

\vspace{-0.5em}
\caption{Retrieval results, ordered from Top-1 to Top-5, for "Which supplier did Wendy's recognize as the most ESG responsible in 2023?". Row 1 is high-compression budget (4,32); Row 2 is ColPali. Red boxes mark the ground-truth page.}
\label{fig:case-study-top5}
\end{figure*}

Figure~\ref{fig:case-study-top5} shows a qualitative example where the highly compressed retriever preserves the strong document match. This example is drawn from the \texttt{esg\_reports\_human\_labeled\_v2} dataset, which pertains to ESG reports within the fast-food industry. The query asks, ``Which supplier did Wendy's recognize as the most ESG responsible in 2023?'' The ground-truth page contains the relevant supplier recognition information. Under the high-compression budget $(4,32)$, the trained PaliGemma retriever ranks this ground-truth page first. In contrast, the ColPali baseline ranks the same page fifth, behind pages that are visually and topically related but do not provide the most direct match to the query. This example illustrates that even with substantial layer and dimension reduction, the trained retriever can keep the key evidence at the top of the ranking, while direct baseline retrieval may assign higher scores to less specific pages. In the appendix~\ref{app:additional-case-studies}, we provide additional examples that demonstrate the behavior of \method, along with their specific rankings in the retrieval task.

\subsection{Ablation Study}
\label{sec:ablation}

\subsubsection{PaliGemma Objective Ablations}

\begin{figure}[t]
\centering
\includegraphics[width=\linewidth]{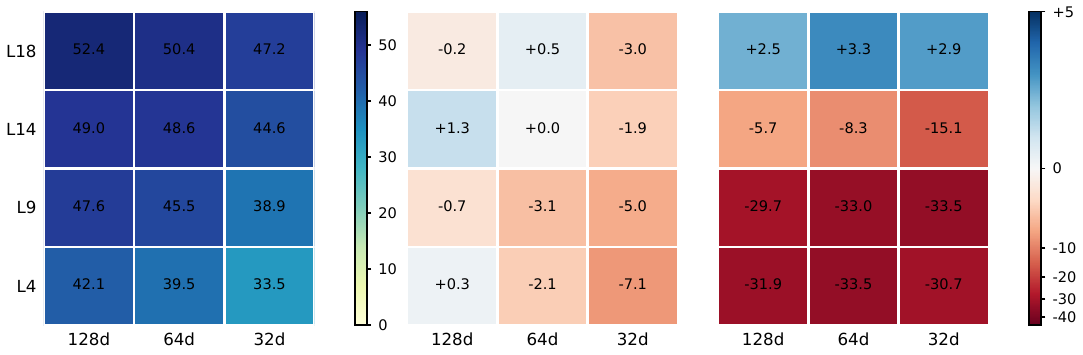}
\vspace{-0.4em}
\begin{tabular}{@{}p{0.28\linewidth}p{0.28\linewidth}p{0.3\linewidth}@{}}
\centering\scriptsize \mbox{(a) Full}&
\centering\scriptsize \mbox{(b) w/o dim MRL--Full} &
\centering\arraybackslash\scriptsize \mbox{(c) w/o layer MRL--Full}
\end{tabular}
\vspace{-0.3em}
\caption{PaliGemma objective ablations over the 2D grid on ViDoRe v2. Panel (a) reports full-objective average nDCG@5 multiplied by $100$; panels (b) and (c) report score differences against the full objective. Rows are layer budgets and columns are prefix dimensions.}
\label{fig:paligemma-ablation}
\vspace{-4mm}
\end{figure}

Figure~\ref{fig:paligemma-ablation} tests whether the two objectives control different parts of the 2D budget space rather than providing redundant supervision. We report this ablation on PaliGemma for both removal variants under the same ViDoRe v2 evaluation protocol used in the main results. The full objective combines the weighted dimension-level and layer-level \method losses. We use \emph{w/o dim MRL} to denote the variant that removes the prefix-dimensional loss and trains the full-width retrieval objective together with layer-wise distillation, so lower-dimensional scores are obtained only by direct truncation. We use \emph{w/o layer MRL} as shorthand for the variant that keeps prefix-dimensional supervision but removes the score-level alignment from the final layer to intermediate layers. This design separates the two intended failure modes: whether retrieval-relevant information is organized into the prefix dimensions, and whether intermediate layers can reproduce final-layer retrieval decisions.

The two removal variants show complementary failure modes. Removing the dimension MRL loss leaves the full-width scores largely intact, but low-dimensional prefixes become less reliable: the final-layer $32$d score drops by $3.0$ points, the $L9$-$D32$ score drops by $5.0$ points, and the $L4$-$D32$ score drops by $7.1$ points. In contrast, removing layer-wise distillation does not damage the final layer in this run, but it severely weakens intermediate-layer retrieval. At middle layer, the drops are $29.7$--$33.5$ points across the three prefix dimensions, and layer $4$ shows similarly large losses. These results support the intended decomposition of \method: prefix-dimensional supervision makes narrow token vectors usable, while score distillation is the main mechanism that makes intermediate layers retrieval-ready under the same \texttt{MaxSim} interface.

\subsubsection{Hyperparameter Analysis}
\label{sec:hparam-analysis}

\begin{figure}[t]
\centering
\includegraphics[width=\linewidth]{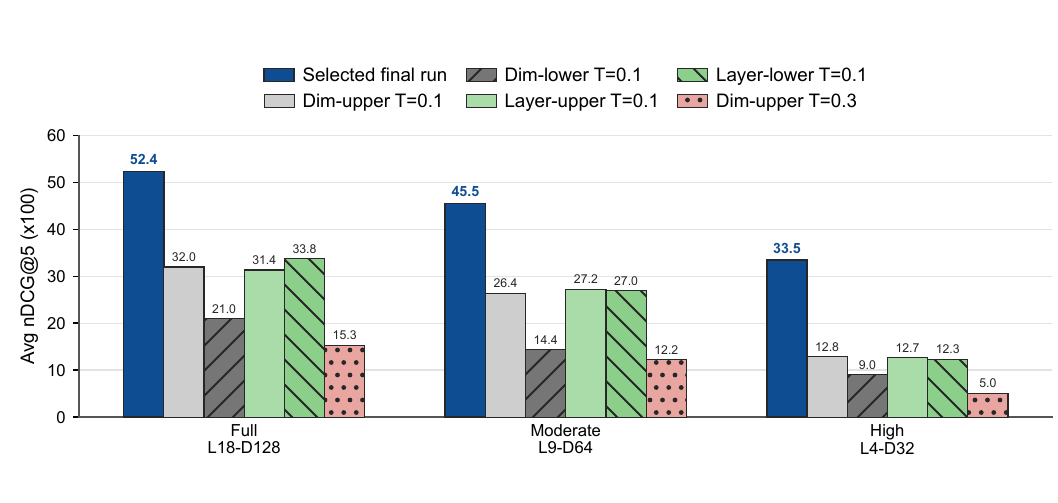}
\caption{PaliGemma hyperparameter comparison on ViDoRe v2. Bars report average nDCG@5 multiplied by $100$ at three selectable budgets.}
\label{fig:hparam-ndcg}
\vspace{-4mm}
\end{figure}

We further compare the selected PaliGemma training configuration with hyperparameter sweep variants---specifically, configurations where dimension weights are biased toward lower or higher dimensions, or layer weights are biased toward shallower layers, or temperature is higher---on the same four ViDoRe v2 datasets. Figure~\ref{fig:hparam-ndcg} reports average nDCG@5 at three representative selectable budgets: full budget ($L18$-$D128$), moderate compression ($L9$-$D64$), and high compression ($L4$-$D32$). The selected configuration achieves the best score at all three budgets, showing that the reported PaliGemma result is not produced by a weak or unstable hyperparameter choice. Appendix~\ref{app:hparam-sensitivity} further reports the complete 2D grid.

The selected PaliGemma run obtains $52.38$, $45.53$, and $33.46$ average nDCG@5 at these three budgets, respectively. In contrast, the second strongest alternative at each budget reaches only $33.75$, $27.24$, and $12.85$, leaving margins of $18.63$, $18.29$, and $20.61$ points. This shows that the current hyperparameter choice is not only better at the full budget, but also preserves a substantially stronger retrieval surface under both moderate and aggressive compression. Appendix~\ref{app:hparam-sensitivity} reports the complete 2D grid, where the selected setting remains stronger across the main 2D budgets while alternatives that over-emphasize one side of the budget space are much weaker overall.

\section{Conclusion}
\label{sec:conclusion}
In this work, we propose \method, a novel 2D training framework designed to overcome the prohibitive storage and computational bottlenecks of multi-vector VDR. 
Unlike existing compression techniques that operate at a fixed representation scale, our approach empowers a single retriever to be naturally budget-elastic along both encoder layer depth and embedding vector width. 
By jointly optimizing dimension-level prefix supervision and layer-wise score distillation, \method allows practitioners to dynamically select a 2D inference budget tailored to specific latency and storage constraints. 
Extensive experiments demonstrate that \method gracefully balances efficiency and accuracy, far better than direct truncation baselines. 
Ultimately, \method establishes a foundational paradigm for resource-constrained deployment, paving the way for the next generation of scalable multimodal retrieval systems.

% \clearpage
\section*{Limitations}
\label{sec:limitations}
While \method successfully establishes a robust budget-elastic paradigm for efficient VDR, there remain a few promising avenues for further exploration:

\begin{itemize}
    \item \textbf{Hardware-aware Deployment Optimization:} Although our framework theoretically supports depth-level elasticity, achieving optimal end-to-end inference speedups across diverse specialized hardware requires bespoke system-level engineering for early-exit execution. We plan to develop custom deployment configurations and integrate them into popular inference engines (\textit{e.g.,} TensorRT or vLLM) to fully actualize these hardware-aware latency gains in industrial settings.

    \item \textbf{Dynamic Budget Allocation:} The current methodology relies on a user-defined, statically selected 2D budget at inference time, rather than dynamically adjusting the computation based on the inherent complexity of individual queries or document pages. In future work, we aim to design an instance-aware routing mechanism that automatically predicts and assigns the minimal sufficient layer-dimension pair for each sample without sacrificing overall accuracy.

    \item \textbf{Extension to Temporal Modalities:} Our empirical scope is currently dedicated to spatial document images, leaving the framework's adaptability to continuous temporal modalities unverified. We intend to expand this 2D Matryoshka paradigm to video-language models to investigate whether temporal frame-level elasticity can be seamlessly incorporated into our joint budget space.
\end{itemize}

\bibliography{mm_matryoshka}

\clearpage
\appendix
\section{Raw Two-Dimensional Compression Results}
\label{app:raw-compression}

Tables~\ref{tab:raw-compression-v1} and~\ref{tab:raw-compression-v2} report the raw nDCG@5 scores. Values are multiplied by $100$. We apply unified dimensions of 128d, 64d, and 32d across all models, while configuring different target layers tailored to each specific backbone.
\begin{table}[t]
\centering
\tiny
\setlength{\tabcolsep}{1.2pt}
\renewcommand{\arraystretch}{0.92}
\resizebox{\columnwidth}{!}{%
\begin{tabular}{ccr*{9}{r}}
\toprule
\multirow{2}{*}{Backbone} & \multirow{2}{*}{Model} & \multirow{2}{*}{Layer} &
\multicolumn{3}{c}{$128$d} & \multicolumn{3}{c}{$64$d} & \multicolumn{3}{c}{$32$d} \\
\cmidrule(lr){4-6}\cmidrule(lr){7-9}\cmidrule(lr){10-12}
& & & R@1 & R@5 & N@5 & R@1 & R@5 & N@5 & R@1 & R@5 & N@5 \\
\midrule
\multirow{10}{*}{PaliGemma} & \multirow{5}{*}{ColPali} & 18 & 70.52 & 84.73 & 78.33 & 68.80 & 82.39 & 76.22 & 61.16 & 76.27 & 69.20 \\
 &  & 14 & 57.67 & 72.61 & 65.93 & 46.81 & 62.15 & 55.17 & 25.14 & 39.02 & 32.72 \\
 &  & 9 & 32.29 & 50.50 & 42.12 & 22.58 & 35.16 & 29.24 & 7.63 & 13.07 & 10.51 \\
 &  & 4 & 24.48 & 41.21 & 33.52 & 15.26 & 24.83 & 20.39 & 6.02 & 11.34 & 8.76 \\
 &  & 2 & 6.95 & 14.65 & 10.89 & 4.21 & 8.16 & 6.31 & 1.89 & 4.20 & 3.13 \\
\cmidrule(lr){2-12}
 & \multirow{5}{*}{+ \method} & 18 & 69.87 & 83.71 & 77.44 & 69.43 & 82.79 & 76.71 & 66.91 & 81.88 & 74.99 \\
 &  & 14 & 70.33 & 83.45 & 77.49 & 68.53 & 82.44 & 76.14 & 65.85 & 81.32 & 74.18 \\
 &  & 9 & 69.46 & 82.13 & 76.43 & 66.87 & 81.18 & 74.80 & 65.52 & 78.89 & 72.70 \\
 &  & 4 & 62.82 & 78.01 & 71.20 & 61.81 & 77.45 & 70.30 & 57.83 & 74.92 & 67.11 \\
 &  & 2 & 37.62 & 57.19 & 47.99 & 34.20 & 53.02 & 44.17 & 29.01 & 46.92 & 38.44 \\
\midrule
\multirow{10}{*}{Qwen2-VL} & \multirow{5}{*}{ColQwen2} & 28 & 75.44 & 89.06 & 82.96 & 75.69 & 88.57 & 82.81 & 69.94 & 84.79 & 78.11 \\
 &  & 21 & 69.41 & 82.87 & 76.75 & 65.01 & 80.03 & 73.33 & 55.45 & 72.61 & 64.60 \\
 &  & 14 & 56.12 & 69.79 & 63.56 & 39.80 & 55.90 & 48.26 & 23.12 & 37.74 & 30.70 \\
 &  & 7 & 62.73 & 76.81 & 70.42 & 53.18 & 70.05 & 62.36 & 40.98 & 56.91 & 49.31 \\
 &  & 4 & 60.36 & 75.66 & 68.96 & 51.27 & 68.72 & 60.78 & 36.22 & 50.64 & 43.93 \\
\cmidrule(lr){2-12}
 & \multirow{5}{*}{+ \method} & 28 & 63.33 & 81.41 & 73.26 & 62.62 & 81.67 & 72.93 & 59.94 & 79.32 & 70.36 \\
 &  & 21 & 62.91 & 80.90 & 72.86 & 62.44 & 80.19 & 72.19 & 59.46 & 78.71 & 69.83 \\
 &  & 14 & 62.66 & 79.68 & 71.96 & 60.21 & 78.38 & 70.17 & 57.52 & 75.01 & 67.07 \\
 &  & 7 & 61.99 & 77.23 & 70.31 & 59.62 & 76.15 & 68.60 & 55.48 & 72.59 & 64.71 \\
 &  & 4 & 57.70 & 74.57 & 66.98 & 55.76 & 73.37 & 65.57 & 51.83 & 69.53 & 61.51 \\
\midrule
\multirow{8}{*}{Qwen2.5-VL} & \multirow{4}{*}{ColQwen2.5} & 36 & 77.10 & 89.06 & 83.62 & 74.35 & 88.01 & 81.94 & 69.36 & 83.69 & 77.13 \\
 &  & 18 & 57.24 & 71.56 & 65.03 & 50.21 & 63.91 & 57.73 & 33.25 & 45.98 & 39.99 \\
 &  & 9 & 57.77 & 71.85 & 65.42 & 46.62 & 61.30 & 54.59 & 24.58 & 36.70 & 30.76 \\
 &  & 6 & 41.57 & 56.14 & 49.35 & 27.86 & 41.59 & 35.19 & 5.98 & 10.90 & 8.44 \\
\cmidrule(lr){2-12}
 & \multirow{4}{*}{+ \method} & 36 & 64.47 & 84.30 & 75.43 & 65.47 & 84.03 & 75.53 & 63.99 & 83.24 & 74.49 \\
 &  & 18 & 67.25 & 84.39 & 76.71 & 66.80 & 83.88 & 76.17 & 63.46 & 82.06 & 73.60 \\
 &  & 9 & 64.62 & 81.39 & 73.97 & 63.31 & 80.06 & 72.72 & 59.60 & 78.22 & 69.91 \\
 &  & 6 & 62.85 & 76.98 & 70.63 & 58.33 & 73.34 & 66.56 & 47.56 & 62.49 & 55.73 \\
\bottomrule
\end{tabular}
}
\caption{Raw ViDoRe v1 results. Values are multiplied by $100$. Under each prefix dimension, the three subcolumns report R@1, R@5, and N@5 (nDCG@5).}
\label{tab:raw-compression-v1}
\end{table}

\begin{table}[t]
\centering
\tiny
\setlength{\tabcolsep}{1.2pt}
\renewcommand{\arraystretch}{0.92}
\resizebox{\columnwidth}{!}{%
\begin{tabular}{ccr*{9}{r}}
\toprule
\multirow{2}{*}{Backbone} & \multirow{2}{*}{Model} & \multirow{2}{*}{Layer} &
\multicolumn{3}{c}{$128$d} & \multicolumn{3}{c}{$64$d} & \multicolumn{3}{c}{$32$d} \\
\cmidrule(lr){4-6}\cmidrule(lr){7-9}\cmidrule(lr){10-12}
& & & R@1 & R@5 & N@5 & R@1 & R@5 & N@5 & R@1 & R@5 & N@5 \\
\midrule
\multirow{10}{*}{PaliGemma} & \multirow{5}{*}{ColPali} & 18 & 49.56 & 79.40 & 52.28 & 45.88 & 77.84 & 49.48 & 37.75 & 67.00 & 38.99 \\
 &  & 14 & 37.15 & 61.40 & 36.79 & 28.05 & 48.95 & 27.58 & 15.71 & 32.02 & 15.81 \\
 &  & 9 & 20.53 & 36.99 & 19.57 & 9.76 & 23.25 & 10.48 & 8.54 & 16.83 & 7.55 \\
 &  & 4 & 15.34 & 27.87 & 13.97 & 8.74 & 18.93 & 8.22 & 6.25 & 12.92 & 5.63 \\
 &  & 2 & 7.23 & 13.35 & 6.78 & 4.91 & 9.73 & 4.09 & 1.91 & 8.09 & 2.60 \\
\cmidrule(lr){2-12}
 & \multirow{5}{*}{+ \method} & 18 & 52.19 & 79.63 & 52.38 & 49.99 & 77.27 & 50.38 & 47.29 & 73.97 & 47.22 \\
 &  & 14 & 47.95 & 76.45 & 48.99 & 48.25 & 75.57 & 48.60 & 44.42 & 72.76 & 44.57 \\
 &  & 9 & 47.81 & 75.53 & 47.64 & 43.75 & 72.64 & 45.53 & 36.52 & 65.63 & 38.93 \\
 &  & 4 & 37.80 & 70.10 & 42.06 & 35.58 & 67.33 & 39.54 & 28.98 & 60.39 & 33.46 \\
 &  & 2 & 16.89 & 35.92 & 18.22 & 12.69 & 31.66 & 15.16 & 10.34 & 27.24 & 12.32 \\
\midrule
\multirow{10}{*}{Qwen2-VL} & \multirow{5}{*}{ColQwen2} & 28 & 52.73 & 81.94 & 53.95 & 51.90 & 78.00 & 50.74 & 40.43 & 70.73 & 42.14 \\
 &  & 21 & 48.75 & 76.62 & 48.53 & 42.37 & 70.91 & 43.02 & 31.71 & 56.62 & 32.22 \\
 &  & 14 & 27.52 & 51.70 & 28.21 & 15.16 & 28.83 & 14.96 & 9.50 & 24.41 & 11.26 \\
 &  & 7 & 33.67 & 63.85 & 36.43 & 24.91 & 49.78 & 26.75 & 17.89 & 34.72 & 18.44 \\
 &  & 4 & 36.23 & 65.28 & 37.55 & 27.61 & 51.27 & 28.34 & 18.57 & 37.73 & 19.69 \\
\cmidrule(lr){2-12}
 & \multirow{5}{*}{+ \method} & 28 & 46.95 & 75.28 & 48.49 & 44.60 & 75.29 & 47.40 & 42.29 & 72.72 & 45.33 \\
 &  & 21 & 43.63 & 74.87 & 47.23 & 43.38 & 72.27 & 45.52 & 40.70 & 71.31 & 43.46 \\
 &  & 14 & 43.99 & 71.91 & 45.38 & 40.65 & 69.16 & 41.90 & 36.18 & 64.45 & 37.88 \\
 &  & 7 & 37.50 & 65.58 & 40.37 & 36.75 & 65.68 & 39.13 & 34.60 & 60.45 & 35.09 \\
 &  & 4 & 39.85 & 69.69 & 41.85 & 36.91 & 64.38 & 38.25 & 34.41 & 54.74 & 32.83 \\
\midrule
\multirow{8}{*}{Qwen2.5-VL} & \multirow{4}{*}{ColQwen2.5} & 36 & 57.28 & 85.55 & 59.12 & 57.25 & 83.86 & 57.19 & 46.16 & 76.01 & 48.08 \\
 &  & 18 & 35.20 & 61.28 & 35.49 & 29.00 & 52.55 & 29.38 & 18.42 & 38.08 & 18.54 \\
 &  & 9 & 29.38 & 58.93 & 31.66 & 20.55 & 47.90 & 23.97 & 10.69 & 24.61 & 12.16 \\
 &  & 6 & 18.98 & 39.58 & 19.73 & 13.05 & 30.96 & 13.90 & 3.80 & 11.85 & 4.16 \\
\cmidrule(lr){2-12}
 & \multirow{4}{*}{+ \method} & 36 & 47.57 & 77.13 & 49.87 & 47.55 & 77.12 & 49.75 & 43.81 & 74.52 & 46.01 \\
 &  & 18 & 48.52 & 77.45 & 50.18 & 47.64 & 73.82 & 47.90 & 42.39 & 69.49 & 42.78 \\
 &  & 9 & 43.19 & 73.94 & 45.16 & 42.67 & 70.29 & 42.52 & 35.43 & 62.47 & 36.36 \\
 &  & 6 & 37.86 & 63.68 & 38.27 & 32.42 & 56.33 & 33.50 & 22.36 & 42.87 & 23.10 \\
\bottomrule
\end{tabular}
}
\caption{Raw ViDoRe v2 results. Values are multiplied by $100$. Under each prefix dimension, the three subcolumns report R@1, R@5, and N@5 (nDCG@5).}
\label{tab:raw-compression-v2}
\end{table}

\section{PaliGemma Efficiency Protocol}
\label{app:efficiency}

We measure PaliGemma efficiency on a single NVIDIA H100 80GB HBM3 GPU with an Intel Xeon Platinum 8468 CPU, PyTorch 2.9.1+cu128, CUDA 12.8, Python 3.10.19, and bf16 precision. The benchmark uses the same four ViDoRe v2 datasets as the main results. Encoding latency is measured with batch size 1. The true early-exit run executes only the selected number of Gemma transformer layers for both query and page encoding, while page encoding still includes the full vision tower. \texttt{MaxSim} scoring uses precomputed query and page embeddings, evaluates 1000-candidate sets, runs 100 measured iterations per dataset and budget, and synchronizes CUDA before and after each timed call. Scoring timing includes host-to-device transfer of the query embedding and host-side score transfer, but excludes VLM encoder forward time, first-stage candidate retrieval, and disk I/O. Table~\ref{tab:paligemma-deployment-efficiency-grid} reports the raw deployment-efficiency grid used to interpret the online and indexing/storage figures. Query and page-indexing latencies are averaged over the four ViDoRe v2 datasets from the true early-exit run; the \texttt{MaxSim} score column reports the corresponding four-dataset 1000-candidate mean latency.

\begin{table*}[t]
\centering
\scriptsize
\setlength{\tabcolsep}{3pt}
\resizebox{\textwidth}{!}{%
\begin{tabular}{cccccccccc}
\toprule
Layer & Dim. & Query & Score@1000 & Online & Online sp. & Index & Index sp. & Index GiB & Storage \\
& & ms/q & ms/q & ms/q & $\times$ & ms/page & $\times$ & & $\times$ \\
\midrule
18 & 128 & 38.33 & 59.48 & 97.81 & 1.00 & 84.66 & 1.00 & 1.117 & 1.0 \\
18 & 64  & 38.33 & 37.44 & 75.77 & 1.29 & 84.66 & 1.00 & 0.558 & 2.0 \\
18 & 32  & 38.33 & 22.60 & 60.93 & 1.61 & 84.66 & 1.00 & 0.279 & 4.0 \\
\midrule
14 & 128 & 27.17 & 59.02 & 86.19 & 1.13 & 69.50 & 1.22 & 1.117 & 1.0 \\
14 & 64  & 27.17 & 36.95 & 64.12 & 1.53 & 69.50 & 1.22 & 0.558 & 2.0 \\
14 & 32  & 27.17 & 23.87 & 51.04 & 1.92 & 69.50 & 1.22 & 0.279 & 4.0 \\
\midrule
9 & 128 & 23.19 & 57.72 & 80.91 & 1.21 & 66.14 & 1.28 & 1.117 & 1.0 \\
9 & 64  & 23.19 & 37.12 & 60.31 & 1.62 & 66.14 & 1.28 & 0.558 & 2.0 \\
9 & 32  & 23.19 & 22.89 & 46.08 & 2.12 & 66.14 & 1.28 & 0.279 & 4.0 \\
\midrule
4 & 128 & 17.49 & 59.50 & 76.99 & 1.27 & 52.24 & 1.62 & 1.117 & 1.0 \\
4 & 64  & 17.49 & 37.47 & 54.96 & 1.78 & 52.24 & 1.62 & 0.558 & 2.0 \\
4 & 32  & 17.49 & 22.66 & 40.15 & 2.44 & 52.24 & 1.62 & 0.279 & 4.0 \\
\bottomrule
\end{tabular}%
}
\caption{Raw PaliGemma deployment-efficiency grid under true early exit. Query and online latencies are in milliseconds per query; Index is page-encoding latency in milliseconds per page. Online latency is Query plus 1000-candidate \texttt{MaxSim} score latency. Speedups are relative to $L18$-$D128$. The true early-exit run currently covers layers 18, 14, 9, and 4.}
\label{tab:paligemma-deployment-efficiency-grid}
\end{table*}

\section{Additional Case Studies}
\label{app:additional-case-studies}

\begin{figure*}[t]
\centering
\includegraphics[width=\textwidth]{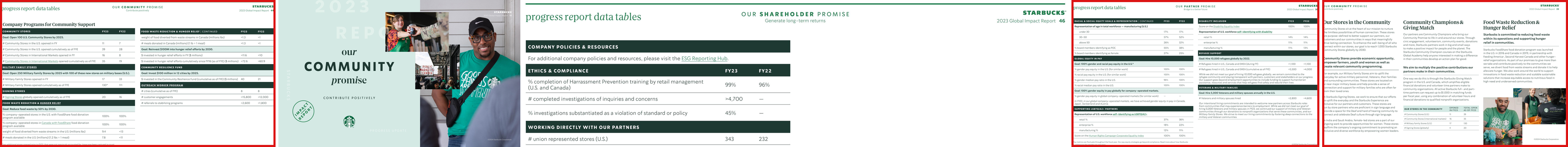}

\vspace{0.35em}
\includegraphics[width=\textwidth]{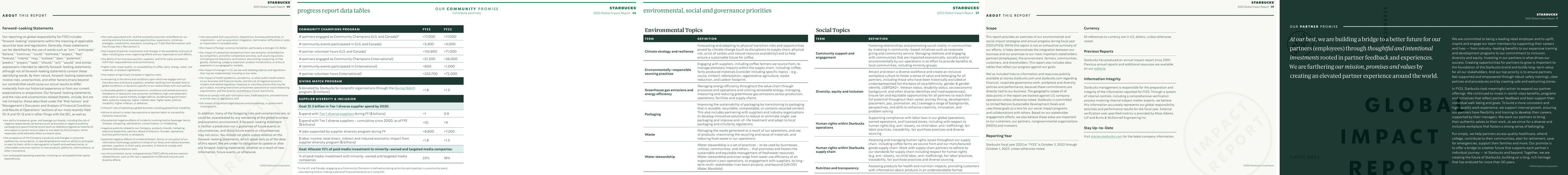}
\caption{Top-5 retrieval results for the query ``What is Starbucks military integration policy?'' Top: ours $(4,32)$, with a gold page at rank 1. Bottom: ColPali, whose first gold page is ranked 7 and therefore does not appear in the displayed top five.}
\label{fig:case-additional-starbucks-policy}
\end{figure*}

\begin{figure*}[t]
\centering
\includegraphics[width=\textwidth]{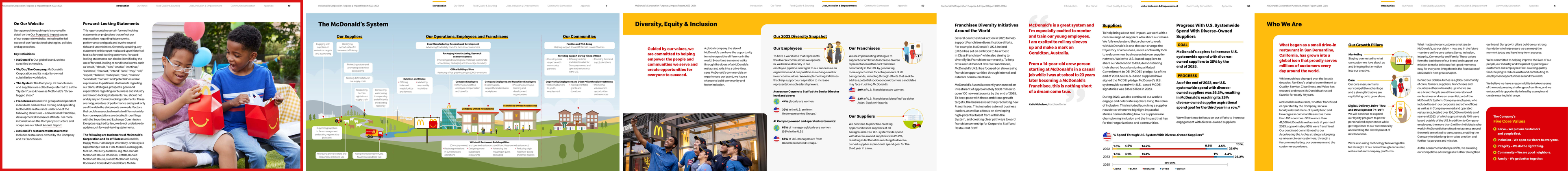}

\vspace{0.35em}
\includegraphics[width=\textwidth]{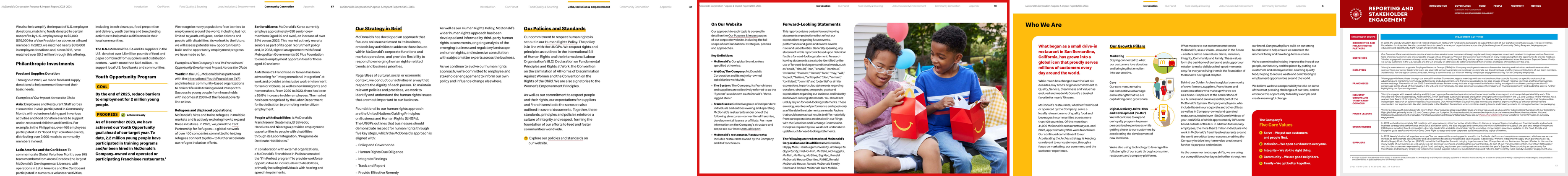}
\caption{Top-5 retrieval results for the query ``What are McDonald's Franchisees?'' Top: ours $(4,32)$, with the gold page at rank 1. Bottom: ColPali, with the gold page at rank 3.}
\label{fig:case-additional-mcdonalds-franchisees}
\end{figure*}

\begin{figure*}[t]
\centering
\includegraphics[width=\textwidth]{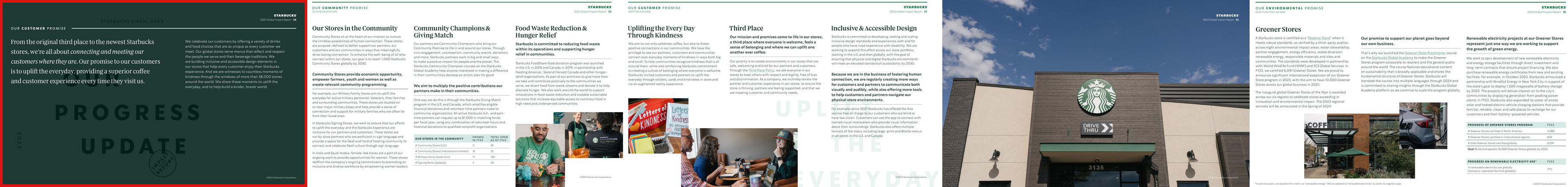}

\vspace{0.35em}
\includegraphics[width=\textwidth]{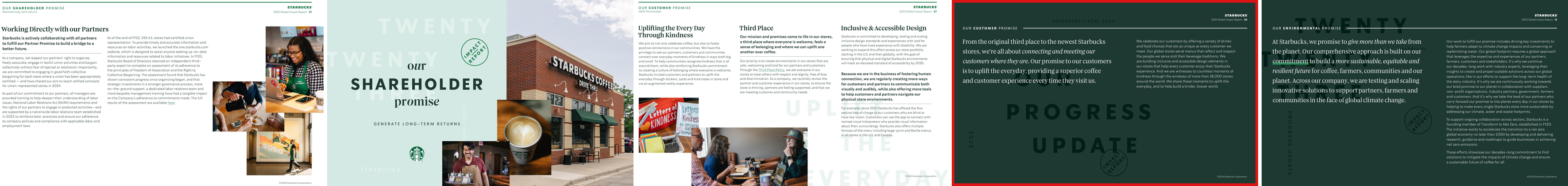}
\caption{Top-5 retrieval results for the query ``How many shops does Starbucks have?'' Top: ours $(4,32)$, with the gold page at rank 1. Bottom: ColPali, with the gold page at rank 4.}
\label{fig:case-additional-starbucks-shops}
\end{figure*}

\begin{figure*}[t]
\centering
\includegraphics[width=\textwidth]{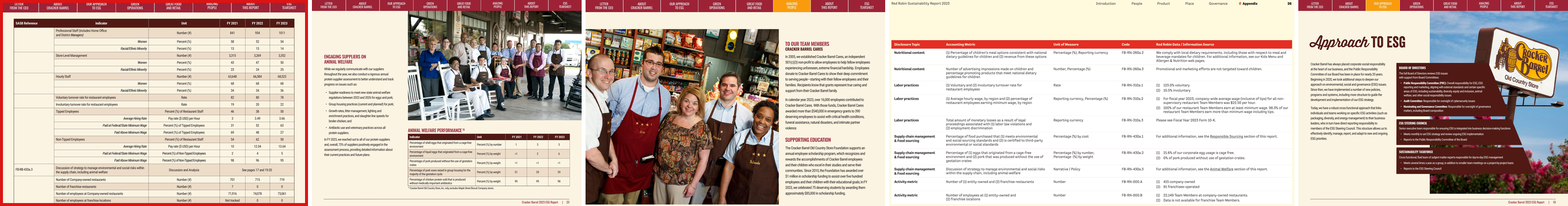}

\vspace{0.35em}
\includegraphics[width=\textwidth]{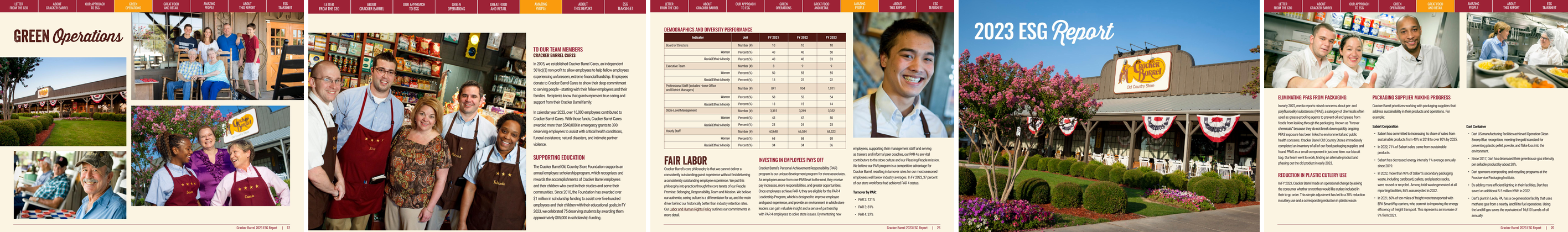}
\caption{Top-5 retrieval results for the query ``How did wages evolve between 2023 and 2021 at Cracker Barrel?'' Top: ours $(4,32)$, with a gold page at rank 1. Bottom: ColPali, whose first gold page is ranked 17 and therefore does not appear in the displayed top five.}
\label{fig:case-additional-cracker-wages}
\end{figure*}

\begin{figure*}[t]
\centering
\includegraphics[width=\textwidth]{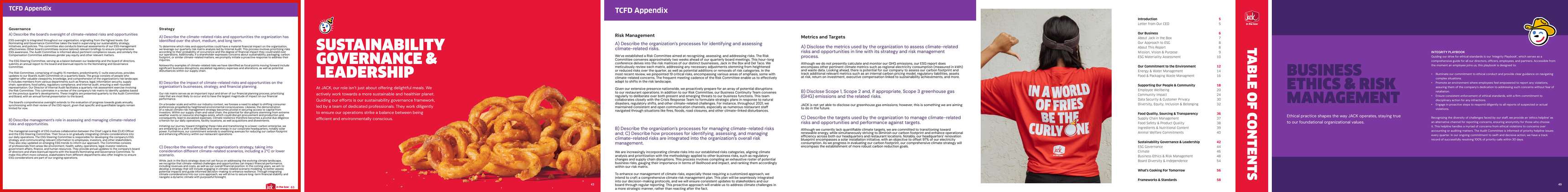}

\vspace{0.35em}
\includegraphics[width=\textwidth]{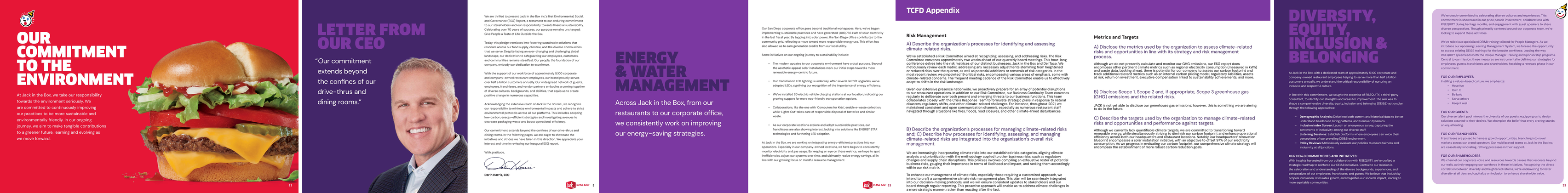}
\caption{Top-5 retrieval results for the query ``Who is responsible for integrating climate consideration into Jack in the Box governance?'' Top: ours $(4,32)$, with a gold page at rank 1. Bottom: ColPali, whose first gold page is ranked 178 and therefore does not appear in the displayed top five.}
\label{fig:case-additional-jack-climate}
\end{figure*}

We additionally examine five queries from the human-labeled ESG reports subset of ViDoRe v2. Each case directly compares the PaliGemma high-compression budget $(4,32)$ with ColPali under the same retrieval setting. Table~\ref{tab:additional-case-ranks} summarizes the rank of the first gold page. In all five selected examples, the compressed retriever places a gold page at rank 1, whereas ColPali ranks the first gold page between rank 3 and rank 178. Figures~\ref{fig:case-additional-starbucks-policy}--\ref{fig:case-additional-jack-climate} show their complete top-5 result lists; red boxes indicate gold pages.

\begin{table*}[t]
\centering
\small
\begin{tabular}{p{0.68\textwidth}cc}
\toprule
Query & ColPali & Ours $(4,32)$ \\
\midrule
What is Starbucks military integration policy? & 7 & \textbf{1} \\
What are McDonald's Franchisees? & 3 & \textbf{1} \\
How many shops does Starbucks have? & 4 & \textbf{1} \\
How did wages evolve between 2023 and 2021 at Cracker Barrel? & 17 & \textbf{1} \\
Who is responsible for integrating climate consideration into Jack in the Box governance? & 178 & \textbf{1} \\
\bottomrule
\end{tabular}
\caption{Additional case studies overview.}
\label{tab:additional-case-ranks}
\end{table*}

\section{Hyperparameter Comparison Details}
\label{app:hparam-sensitivity}
Table~\ref{tab:hparam-full-grid} expands the PaliGemma hyperparameter sweep to the full 2D grid. All numbers are four-dataset ViDoRe v2 averages; each cell reports retention relative to that sweep setting's own full budget, with the absolute nDCG@5 value multiplied by $100$ in parentheses.
\begin{table*}[t]
    \centering
    \scriptsize
    \setlength{\tabcolsep}{2.5pt}
    \resizebox{\textwidth}{!}{%
    \begin{tabular}{lccccc}
    \toprule
    Budget & Dim-upper, $T=0.1$ & Dim-lower, $T=0.1$ & Layer-upper, $T=0.1$ & Layer-lower, $T=0.1$ & Dim-upper, $T=0.3$ \\
    \midrule
    $L18$-$D128$ & 100.0 (32.00) & 100.0 (21.01) & 100.0 (31.37) & 100.0 (33.75) & 100.0 (15.26) \\
    $L18$-$D64$ & 96.1 (30.74) & 89.8 (18.87) & 97.8 (30.67) & 93.2 (31.45) & 98.8 (15.08) \\
    $L18$-$D32$ & 85.7 (27.41) & 83.2 (17.49) & 84.5 (26.52) & 81.0 (27.35) & 93.0 (14.19) \\
    \midrule
    $L14$-$D128$ & 95.3 (30.50) & 104.1 (21.87) & 99.6 (31.24) & 89.9 (30.33) & 117.6 (17.94) \\
    $L14$-$D64$ & 88.5 (28.33) & 97.3 (20.45) & 95.5 (29.97) & 85.4 (28.81) & 116.7 (17.81) \\
    $L14$-$D32$ & 84.1 (26.90) & 87.9 (18.47) & 89.0 (27.93) & 74.4 (25.10) & 102.2 (15.60) \\
    \midrule
    $L9$-$D128$ & 95.4 (30.54) & 81.4 (17.11) & 96.1 (30.14) & 90.8 (30.65) & 93.6 (14.28) \\
    $L9$-$D64$ & 82.4 (26.36) & 68.5 (14.39) & 86.8 (27.24) & 79.9 (26.96) & 80.0 (12.21) \\
    $L9$-$D32$ & 66.7 (21.33) & 54.5 (11.44) & 70.3 (22.04) & 64.9 (21.89) & 63.7 (9.72) \\
    \midrule
    $L4$-$D128$ & 62.8 (20.11) & 58.5 (12.30) & 64.0 (20.09) & 58.0 (19.58) & 37.7 (5.75) \\
    $L4$-$D64$ & 50.6 (16.18) & 49.3 (10.36) & 51.5 (16.15) & 48.8 (16.48) & 34.7 (5.29) \\
    $L4$-$D32$ & 40.2 (12.85) & 42.8 (8.99) & 40.4 (12.66) & 36.4 (12.28) & 33.1 (5.05) \\
    \midrule
    $L2$-$D128$ & 35.9 (11.49) & 30.5 (6.41) & 31.4 (9.84) & 35.8 (12.09) & 21.8 (3.33) \\
    $L2$-$D64$ & 25.9 (8.29) & 28.7 (6.04) & 24.1 (7.55) & 26.4 (8.90) & 22.2 (3.39) \\
    $L2$-$D32$ & 22.6 (7.23) & 28.9 (6.08) & 21.8 (6.84) & 24.1 (8.15) & 20.4 (3.11) \\
    \bottomrule
    \end{tabular}%
    }
    \caption{Full PaliGemma hyperparameter analysis grid on the four ViDoRe v2 datasets. Each cell reports retention relative to the same setting's $L18$-$D128$ score, with absolute nDCG@5 in parentheses.}
    \label{tab:hparam-full-grid}
    \end{table*}

\section{Reproducibility Details}
\label{app:reproducibility}

\begin{table*}[t]
\centering
\small
\begin{tabular}{llll}
\toprule
Role & Model & Size & Hugging Face repository \\
\midrule
\multirow{3}{*}{Backbone}  & PaliGemma & 3B & \url{https://huggingface.co/google/paligemma-3b-mix-448} \\
 & Qwen2-VL & 2B & \url{https://huggingface.co/Qwen/Qwen2-VL-2B-Instruct} \\
 & Qwen2.5-VL & 3B & \url{https://huggingface.co/Qwen/Qwen2.5-VL-3B-Instruct} \\
\midrule
\multirow{3}{*}{Baseline}  & ColPali & 3B & \url{https://huggingface.co/vidore/colpali-v1.3} \\
 & ColQwen2 & 2B & \url{https://huggingface.co/vidore/colqwen2-v1.0} \\
 & ColQwen2.5 & 3B & \url{https://huggingface.co/vidore/colqwen2.5-v0.2} \\
\bottomrule
\end{tabular}
\caption{Backbone and baseline model repositories used in the experiments.}
\label{tab:model-details}
\end{table*}

\paragraph{Evaluation logs and checkpoints.}
\paragraph{Models and checkpoints.}
  We train one \method checkpoint for each backbone family: PaliGemma, Qwen2-VL, and Qwen2.5-VL. Each trained retriever is evaluated against the corresponding public late-interaction baseline: ColPali, ColQwen2, and ColQwen2.5. All reported compression results are obtained from the same checkpoint per backbone by selecting different 2D budgets at inference time, without training separate models for each budget.

\paragraph{Datasets and metric scale.}
The ViDoRe v1 averages use ten datasets: \path{docvqa_test_subsampled}, \path{infovqa_test_subsampled}, \path{tabfquad_test_subsampled}, \path{arxivqa_test_subsampled}, \path{tatdqa_test}, \path{shiftproject_test}, \path{syntheticDocQA_artificial_intelligence_test}, \path{syntheticDocQA_energy_test}, \path{syntheticDocQA_government_reports_test}, and \path{syntheticDocQA_healthcare_industry_test}. The ViDoRe v2 averages use four datasets: \path{esg_reports_v2}, \path{biomedical_lectures_v2}, \path{economics_reports_v2}, and \path{esg_reports_human_labeled_v2}. Table values report nDCG@5 multiplied by $100$. The Qwen2-VL ablation table uses a common three-dataset ViDoRe v2 subset for which all ablation variants completed: ESG reports, biomedical lectures, and human-labeled ESG reports.

\paragraph{Layer and dimension probes.}
The main compression-retention tables report three layer budgets for each backbone: full, middle, and shallow. These correspond to $\{18,9,4,2\}$ for PaliGemma, $\{28,14,7\}$ for Qwen2-VL, and $\{36,18,9\}$ for Qwen2.5-VL. All three backbone families use the common prefix dimensions $\{128,64,32\}$.

\paragraph{Baseline truncation.}
For public baselines and trained models, lower-dimensional evaluation uses the same protocol: slice the prefix dimensions of every query and document token vector, re-normalize the sliced vectors, and score with the same late-interaction \texttt{MaxSim} function. The reported prefix dimensions are shared across backbones, so the compression-retention comparison uses the same $\{128,64,32\}$ dimension grid for PaliGemma, Qwen2-VL, and Qwen2.5-VL.

\end{document}

%% file: packages.tex
\usepackage{graphicx}
\usepackage{subcaption}
\usepackage{multirow}

\usepackage{amsmath}

\usepackage{paralist}

\usepackage[capitalize]{cleveref}
\crefname{section}{Sec.}{Secs.}
\Crefname{section}{Section}{Sections}
\Crefname{table}{Table}{Tables}
\crefname{table}{Tab.}{Tabs.}

\usepackage{microtype}
\usepackage{booktabs} % 

\definecolor{darkgreen}{rgb}{0.0, 0.5, 0.0} 

\usepackage[dvipsnames,table]{xcolor}

\usepackage{tabularx}
\usepackage{multicol}
\definecolor{myblue}{RGB}{235,235,250}

\usepackage{xspace}

\usepackage{pifont}
\usepackage{latexsym}
\usepackage{inconsolata}
\usepackage{arydshln}
\usepackage{enumitem}
\usepackage{wrapfig}
\usepackage{array}
\usepackage{epigraph}

\definecolor{lightpink}{RGB}{204, 231, 207} %lightgreen tabcolor3
\definecolor{lightblue}{RGB}{210, 220, 250} %lightblue tabcolor5
\definecolor{lightgray}{RGB}{237, 237, 237} %lightblue tabcolor5

\definecolor{superlightred}{rgb}{0.99, 0.92, 0.92}
\definecolor{darkgreen}{RGB}{50,100,0}
\definecolor{darkred}{RGB}{200, 0, 0}

\usepackage{makecell}
\usepackage{colortbl}

\usepackage[ruled,vlined]{algorithm2e}

\usepackage{hyperref}
\usepackage{url}
\usepackage{hyperref}
\usepackage[most]{tcolorbox}
\usepackage{wrapfig}
\usepackage{graphicx,xcolor,float}
\usepackage{subcaption}
\usepackage{threeparttable}
\usepackage{pifont}
\usepackage{colortbl}
\usepackage{color}
\usepackage{multirow}
\usepackage{tabularx}
\usepackage{float}
\usepackage{graphicx}
\usepackage{booktabs}
\usepackage{arydshln}
\usepackage{enumitem}
\usepackage{wrapfig}
\usepackage{caption}
\usepackage{graphicx}
\usepackage{makecell}
\usepackage{float} 
\usepackage{makecell}
\usepackage{tabularx}
\usepackage{amssymb,mathrsfs,amsmath}
\usepackage{pifont}
\usepackage[export]{adjustbox}
\usepackage{xcolor}
\usepackage{xspace}

\hypersetup{
    colorlinks=true,
    linkcolor=red,
    citecolor=cyan,
    filecolor=magenta,      
    urlcolor=magenta,
    }

\definecolor{bittersweet}{rgb}{1.0, 0.44, 0.37}
\definecolor{mygreen}{rgb}{0.29, 0.7, 0.48}
\definecolor{my_green}{RGB}{51,102,0}
\definecolor{my_yellow}{RGB}{255,165,0}
\definecolor{my_red}{RGB}{204, 0, 0}
\definecolor{my_orange}{RGB}{232, 132, 23}

\usepackage{pifont}% http://ctan.org/pkg/pifont for xmark and cmark
\definecolor{mygray}{gray}{0.4}
\hypersetup{
    colorlinks=true,
    linkcolor=red,
    citecolor=cyan,
    filecolor=magenta,      
    urlcolor=magenta,
    }
\usepackage{xcolor} % 如果你想为雪花上色
\usepackage{xcolor}
\usepackage[font=small,labelfont=bf]{caption}  % set global caption size
\usepackage{makecell}
\usepackage{tabulary}
\definecolor{ada_green}{rgb}{0,205,205}
\definecolor{glt_red}{rgb}{109,205,255}

\definecolor{backred}{RGB}{255, 190, 190}
\definecolor{backblue}{RGB}{210, 230, 250}
\definecolor{backgrey}{RGB}{220, 220, 220}
\definecolor{green(pigment)}{rgb}{0.0, 0.65, 0.31}